\documentclass[11pt, a4paper]{article}
\usepackage{amsbsy, graphicx, amssymb, amsmath, amsthm, amsfonts, mathrsfs, color, subfig, anysize}
\usepackage[colorlinks, citecolor=blue, urlcolor=blue, linkcolor=black, breaklinks]{hyperref}
\usepackage{hyperref}
\hypersetup{
    pdfborder   ={0 0 0},
    pdfauthor   =Dr. Stefan Hummelsheim,
    pdftitle   =Lasso and equivalent quadratic penalized models,
    pdfsubject =Lasso and equivalent quadratic penalized models,
    pdfpagemode=UseThumbs, 
    pdfkeywords= {Linear regression}; {Penalized regression}; {LASSO}; {Modelselection} ; {Regularization}; {Sparse solution}; 
}

\usepackage[utf8]{inputenc}
\usepackage[english]{babel} 

\marginsize{15mm}{15mm}{15mm}{15mm}
\DeclareMathOperator*{\argmin}{argmin}
\DeclareMathOperator*{\subto}{\quad s.t. \,}
\newcommand{\diag}[1]{\mbox{diag}(#1)}

\begin{document}
\title{Lasso and equivalent quadratic penalized regression models}
\author{Dr. Stefan Hummelsheim\thanks{\copyright \, by the author}}
\date{Version 1.0  Dec 29, 2013}
\maketitle
\thispagestyle{empty}

%\IfFileExists{./workL1L2.tex}{\tableofcontents}{}

\abstract{
The least absolute shrinkage and selection operator (lasso) and ridge regression produce usually
different estimates although input, loss function and parameterization of the penalty are identical.
In this paper we look for ridge and lasso models with identical solution set. 

It turns out, that the lasso model with shrink vector $\lambda$ and a quadratic penalized model with
shrink matrix as outer product of $\lambda$ with itself are equivalent,
in the sense that they have equal solutions. To achieve this, we have to restrict the estimates to be
positive. This doesn't limit the area of application since we can decompose every estimate 
in a positive and negative part. The resulting problem can be solved with a non negative least 
square algorithm and may benefit from algorithms with high numerically accuracy.
This model can also deal with mixtures of ridge and lasso penalties like the elastic net, 
leading to a continuous solution path as a function of the mixture proportions.

Beside this quadratic penalized model, an augmented regression model with positive bounded estimates is
developed which is also equivalent to the lasso model, but is probably faster to solve.

}

\section{Introduction}
More than 40 years ago \cite{Hoe70} introduces ridge regression to overcome problems with
multicollinearity. In the last decades, 
lasso \cite{HTF09} and derivatives like the 
generalized lasso \cite{TS11}, 
the elastic net \cite{FHT10} or
the adaptive lasso \cite{Zu07} 
has become common tools in 
regression analysis for estimating sparse coefficient vectors e.g. with only few non zeros 
thus leads to subset or model selection. Popular other models with comparable objective are  
the Dantzig selector \cite{CT07}, the SCAD \cite{Fan01} or iterated reweighed methods 
as shown in \cite{CWB08}.

Although the objective of the lasso and ridge approach is very different, they have several points
in common:

\begin{itemize} 
\item Both methods can deal with multicollinearity and the case of fat design matrices 
(more columns than rows) of the independent variables do to the shrinkage.
\item Both methods are summarized in literature under the topic regularization and need 
tuning parameters say $\lambda$ which leads to a path of estimates.
\end{itemize}

The key difference between these methods is the functional form of the penalty. Roughly, 
ridge penalty is based on square values whereas lasso penalty is based on absolute values 
of the estimates (Fig. \ref{fig:lasso2ridge}). 
\clearpage

\begin{figure}[h]
\label{fig:lasso2ridge}
\rule{0.10\textwidth}{0mm}
\begin{minipage}{.80\textwidth}\noindent
\includegraphics[width=.80\textwidth]{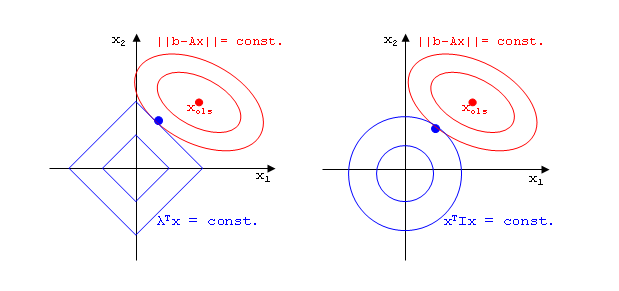}
\caption{Contours of lasso and ridge regression in the space $(x_1,x_2)$ of estimates \label{fig:lasso2ridge}\\
{\small Schematically view of loss (red) and penalties (blue) for lasso (left) and ridge regression (right). 
The red point is the ordinary least square solution, the blue point marks the optimal penalized estimate.}}
\end{minipage}
\end{figure}

This is often graphically illustrated as  rectangle versus ball shape (Fig. \ref{fig:lasso2ridge}) for
the penalty function.

From this point of view both approaches seem to be incompatible. But this is 
not the case and in the following section, we develop pure quadratic models 
-- not $(l_1,l_2)$ mixtures like the elastic net \cite{FHT10} -- with rectangular 
contour shapes as illustrated in Fig. \ref{fig:lasso2ridge}.

\section{Lasso and related regression problems}\label{sec:L2L1}
Given a Matrix $A \in \mathbb{R}^{m\times n}$ and a vector $b \in \mathbb{R}^{m}$ the lasso \cite{HTF09} 
approach looks for a sparse $x \in \mathbb{R}^{n}$ solution which minimizes $||b - Ax||^2_2$.
This is achieved by solving $\mbox{min}_x ||b - Ax||^2_2 +\lambda^T |x|$ with regularization 
parameter $\lambda \in \mathbb{R}^{n}_{+}$ where $|x|= (|x_1|, \ldots, |x_n|)^T$.
This model is equivalent to the generalized lasso \cite{FHT10} with penalty function $\lambda_1 |Dx|$ where $\lambda_1=1, D=\diag{\lambda}$.

In the following derivations, we use this setting and give in the Appendix \label{appendix} a small 
gams \cite{Gams} programm which will solve (for sample data) the presented Models.

\subsection{Lasso with unbounded and bounded $x$}

We start with introducing the definition of the lasso penalized model in equation 
\ref{eq:xfree1}\footnote{$\subto$ = subject to, superscript $(..)^T$ =transpose of $(..)$, $I$= identity Matrix, $\diag v$= diagonal Matrix from vector v}. 
We minimize the sum of the squared residuals $\epsilon^T \epsilon$ -- defined by equation $ b = A x +I\epsilon$ -- 
plus the strictly positive lasso penalty sum $\lambda^T |x|$.

\begin{subequations}\label{eq:xfree}
\begin{align}
 x_{l} & =  \argmin_{x}  \left\{ \frac{1}{2}\epsilon^T \epsilon + \lambda^T |x| \subto b = A x +I\epsilon \right\} \label{eq:xfree1} \\
 x_{l} & =  \argmin_{x}  \left\{ \frac{1}{2}\epsilon^T \epsilon + \lambda^T (x^{+} + x^{-}) \subto b = A x +I\epsilon,\, x = x^{+} - x^{-}, \, x^{+} \ge 0,  x^{-} \ge 0  \right\} \label{eq:xfree2} 
\end{align}
\end{subequations}

Obviously, problem \ref{eq:xfree1} and \ref{eq:xfree2} have the same solution,
since every $x \in \mathbb{R}^{n}$ can be decomposed in a positive $x^{+}$ and
negative part $x^{-}$.
The second model is a continuous version of the first. In the case of mixed
coefficients signs for $x$, this yields formally a model with twice columns. 
But this need not be a disadvantage:

\begin{itemize} 
\item In many cases, where $x$ is restricted in the sign by the underlying real
problem this is no disadvantage and is intended.
\item From a algorithmic point of view it is often not obligatory to use storage
for two A-Matrices (one for each sign) nor to calculate for each sign the
gradients e.g. $A^T \epsilon$ separately.
If the interface to an appropriate solver e.g. IPOPT \cite{wwwIPOPT} or NNLS
\cite{LH74} is not too tight or monolithic other strategies are obvious.
\end{itemize}

The same decomposition pattern can be applied to the residual definition ($\epsilon^{+},\epsilon^{-}$) 
and loss function yielding the least absolute deviation lasso \cite{WLJ06} 
which can be solved with a pure $l_1$-estimator algorithm.\\
	
This derivation \ref{eq:xfree} shows that for every lasso Problem there exists a positive 
constrained model with equal solution set. The benefit of this derivation is, that we get rid of 
3 quadrants of the cartesian coordinate system in \ref{fig:lasso2ridge}.

\subsection{Non negative lasso}
In the following sections we focus on the sign constrained Model, see also \cite{MH12}.

\begin{subequations}\label{eq:nnl}
\begin{align}
 x_{nnl} & = \argmin_{x}  \left\{ \frac{1}{2}\epsilon^T \epsilon + \lambda^T x \subto   b = A x +I\epsilon, x \ge 0 \right\} \label{eq:nnl1} 
\end{align}
\end{subequations}

The Karush–-Kuhn–-Tucker conditions  \cite[p.157]{Do90} for the solution are:
\begin{eqnarray*}
\left[\begin{array}{ccc}
A    & I    & 0\\
0    & A^T  & I\\
\end{array}\right]
\left[\begin{array}{c}
x        \\
\epsilon \\
\mu \\
\end{array}\right]=
\left[\begin{array}{c}
b        \\
\lambda \\
\end{array}\right], \quad x \ge 0,   \,\mu \ge 0,  \,x^T \mu=0 
\end{eqnarray*}

where $\mu$ is the Lagrange multiplier for the  positive constraint $x \ge 0$. 
Note, that for all nonzero x in the solution, the equations $A^T \epsilon = \lambda, \mu=0$ hold. 
This relation is extensively used in several active set algorithms for the lasso problem \cite[Eq. 4]{FHT10}.

\subsection{Augmented regression}
The idea in this section is, to incorporate the penalty in the loss function as one 
additional row of A and b:

\begin{subequations}\label{eq:annl}
\begin{align}
x_{annl} & = \argmin_{x}  \left\{ \frac{1}{2}\tilde{\epsilon}^T \tilde{\epsilon} \subto \tilde{b} = \tilde{A} x +I\tilde{\epsilon}, x \ge 0 \right\} \quad \mbox{with }
\tilde{A} =
\left[\begin{array}{c}
A        \\
\lambda^T\\
\end{array}\right]
,
\tilde{b} =
\left[\begin{array}{c}
b        \\
0\\
\end{array}\right] 
\begin{array}{c}
{\scriptstyle 1, \ldots, m}    \\
{\scriptstyle m+1}\\
\end{array}
\label{eq:annl1} 
\end{align}
\end{subequations}

Notice that $\tilde{A}^T \tilde{\epsilon} = A^T \epsilon +\tilde{\epsilon}_{m+1} \lambda$ 
where $\tilde{\epsilon}_{m+1}$ is the residual for the additional equation ${m+1}$ of the
augmented system. In comparison to the analogous condition for \ref{eq:nnl1}, the
difference is only in the scaling of the penalty by the scalar $\tilde{\epsilon}_{m+1}$. 
But this scaling is for each column of A the same. If we solve \ref{eq:annl1} 
and define $\lambda_1=|0 - \lambda^T x_{annl}|\lambda$ then model \ref{eq:nnl1} 
has for this $\lambda_1$ the same $x$ solution.

\begin{itemize} 
\item The solution to this model can be calculated with e.g. non negative least square \cite{LH74},
	interior point algorithm, or other solver for constrained quadratic programming \cite{Gams}. 
	Because this model is -- except of $x \ge 0$ -- a pure quadratic model, QR decomposition 
	\cite[p. 246]{Golub13} is a stable choice for solving this problem in the case of highly 
	dependent A-columns.

\item To get the solution path $x=x(\lambda)$, increase or decrease the value of the additional
	  $\tilde{b}$ in \ref{eq:annl1}. In case of $QR$ decomposition of $\tilde{A}$,
	  this can be easily done with changes in the right hand side vector $Q^T \tilde{b}$. 
\end{itemize}

In contrast to this model, the augmented model 
$\displaystyle x^T \left[ {A \atop \diag{\lambda}} \right] x= x^T A x + x^T \diag{\lambda} x$ 
results in ridge regression \cite{Hoe70} with Tikhonov matrix $\sqrt{\diag{\lambda}}$.\\

We are now able, to calculate with a pure quadratic model for the underlying lasso model 
the solution vector $x$.

\subsection{Associated quadratic penalty}
The idea is here to extract the implicit penalty from the augmented model \ref{eq:annl1}:

\begin{subequations}\label{eq:qnnl}
\begin{align}
x_{qnnl} &= \argmin_{x}  \left\{ \frac{1}{2}\epsilon^T \epsilon + \frac{1}{2} x^T C x \subto b = A x +I\epsilon, x \ge 0 \right\} \quad \mbox{with } C = \lambda \lambda^T  \label{eq:qnnl1} 
\end{align}
\end{subequations}

This problem has the same solution as \ref{eq:nnl1} and \ref{eq:annl1}. 

\begin{figure}[h]
\centering
\hspace{2mm}
\subfloat[t][ridge, $\alpha=0$]              {\includegraphics[width=0.20\textwidth]{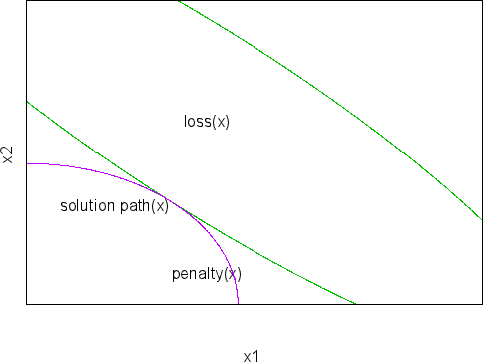}}\hspace{1mm}
\subfloat[t][major axis inc., $\alpha+\!\!+$]{\includegraphics[width=0.20\textwidth]{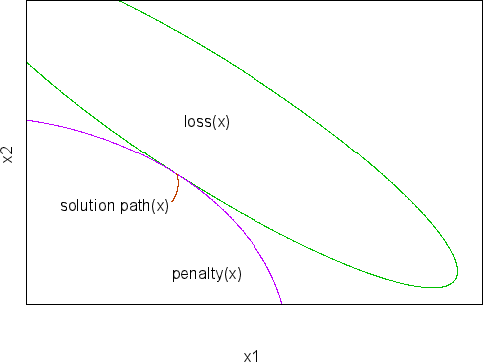}}\hspace{1mm}
\subfloat[t][major axis inc., $\alpha+\!\!+$]{\includegraphics[width=0.20\textwidth]{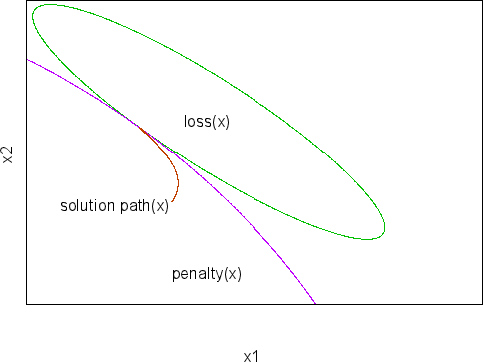}}\hspace{1mm}
\subfloat[t][equivalent quadratic model for lasso,  $\alpha=1$]         {\includegraphics[width=0.20\textwidth]{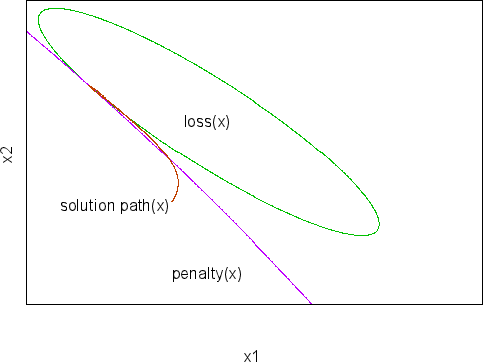}}\\
\caption{Moving from ridge to lasso--regression in solution space $x$}
\label{fig:lossAndPenalty}
\rule{0.10\textwidth}{0mm}
\parbox[t]{0.80\textwidth}{\small where: loss function (green), penalty function (pink), 
solution path (brown) for $\alpha \in [0,1]$. Model for this plot:\\
 $(x-\tilde{x})^T{ 1.0 \, 0.7 \choose 0.7 \, 1.0} (x-\tilde{x}) +x^T\left( (1-\alpha) { 1 \, 0 \choose 0 \, 1}+ \alpha \lambda \lambda^T \right) x$, 
$\lambda_i=1/\tilde{x}_i$, $\tilde{x}={ 2 \choose 3}$\\
}
\end{figure}

Consider the set of x values $\{ x \in \mathbb{R}^{n}_+ | \lambda^T x = c\}$ for 
which the lasso penalty will have the value $c$. On this set, the quadratic penalty
$x^T C x= (\lambda^T x)^T (\lambda^T x) = c^2$ is also constant, but with squared value. 
To see this, imagine the case $x \in \mathbb{R}^2_+$. In Fig. \ref{fig:lossAndPenalty} 
we start in the left panel with traditional ridge penalty matrix $I$ and 
increase the dominance of the major axes of the ellipsoid, until it coincides with the lasso model.

The partial derivative of the penalty $\partial \,(0.5 x^TCx) /
\partial x = \lambda (\lambda^T x) = s(x) \lambda $ is for each
x a multiple $s(x) \in \mathbb{R}$ of the $\lambda$-vector
so that this penalty contributes in each component $n$ similar
to the objective gradient like the lasso penalty.

Furthermore we can build the product of this regularization $CC = C^TC = s C$ and see
that -- up to a scaling factor $s=\lambda^T\lambda$ -- C is a projection matrix.

The contours in the $x$-space of $x^T C x = c$ and $\lambda^T x =c$ 
are in both cases straight lines with the same slope because
the ellipsoid of $x^T C x = c$ has only one principal component.

In conjunction with the decomposition of $x = x^{+} - x^{-}$ in
equation \ref{eq:xfree2} it is easily seen, that it is possible
to construct for every lasso problem a related quadratic penalty
model. The Tikhonov matrix in \cite[pp. 809]{recipes} for the penalized
model $||b-Ax||^2+ ||B x ||^2$ is just $B=\lambda^T$.

Since the quadratic penalty is a strictly monotone
transformation $c^2$ of the lasso penalty this equivalence will
also hold for other loss function e.g. $|b-Ax|$ or Huber
loss \cite{HuberLoss}.\\

There are several computational aspects to mention: 

\begin{itemize} 
\item  In comparison to ''iterated Ridge Regression'' \cite[p.7]{MS05}
this approach does not need reweighting iterations for computing the lasso
solution.

\item If the signs of the $x_j$ to this problem are known, multiplying
the A-columns $a_j$ with $\mbox{sign}(x_j)$ gives a non
restricted regularized least square problem,
for which the usual relations for ridge regression hold e.g. 
$x=(A^T A +C)^{-1}A^T b$ depends linear on b.

\item We are now able to calculate with numerically very stable
algorithms like SVD or QR \cite{Golub13}
solutions for the underlying lasso model and compare this with
solutions of other programs/algorithm like LARS \cite{EHJ04},
interior point methods \cite{wwwIPOPT}, or coordinate descent algorithms.
\end{itemize}

\section{Discussion}

Two of the above models are relevant for practical applications and are
subject for further developments.\\

The Model \ref{eq:annl1} which incorporates the penalty in the objective is
a pure NNLS problem. Its advantage is that it can be solved via numerically
stable QR or SVD algorithms. In future work it will be interesting to
compare model results of traditional lasso models and algorithms -- like the
glmnet-library for R -- with the QR or SVD algorithms. In the case of high
dimensional and dependent A-columns this could result not only in a
quantitative differences in the optimizing $x$ vector, but also in
qualitative different $x(\lambda)$-paths in which other variables have been
identified to be nonzero.\\

The Model \ref{eq:qnnl1} introduces a quadratic penalty as the outer product
of the lasso shrink vector with itself. In conjunction with the $x = x^{+} -
x^{-}$ decomposition we described above a simple way of how to bridge the
$l_1$ regularization of the lasso with the $l_2$ regularization of the ridge
regression. Probably many results for ridge and lasso regression are now
interchangeable, provided that the assumptions for the penalty are not to
strong. This is for instance the case if the ridge penalty matrix is assumed
to have full rank.

\appendix
\section{Appendix}\label{appendix}

The following gams \cite{Gams} source solves\footnote{for solving look at \url{http://www.neos-server.org/neos/solvers/index.html} 
and choose a nlp solver which supports gams} for a random data set the least square problem, the lasso, the non negative lasso, the non 
negative augmented regression and the non negative ridge regression with the penalties of section \ref{sec:L2L1}.
\begin{tiny}
\begin{verbatim} 	
Set     sx / xNoShrink, xlasso, xL2Shrink, xL1Shrink, xRidge, xIni /,
	allrows       / r1 * r10 /, rows(allrows) / r1 * r9 /, shrinkrows(allrows) / r10 /,
	cols          / c1 * c7 /;
Alias (cols, ccols);

Parameter Data(allRows,*), lambda(cols), xSol(*,*), RidgeMatrix(cols,cols), shrinkScale;
xSol(cols,'xIni')=ord(cols);
lambda(cols) = 0.5; 
RidgeMatrix(cols,ccols)=lambda(cols)*lambda(ccols);

Data(      rows,cols)=uniform(0,1);
Data(shrinkrows,cols)=lambda(cols);
Data(rows,       'b')=Sum(cols, Data(rows,cols)*xSol(cols,'xIni'));
Data(rows,       'b')=Data(rows,'b')*uniform(0.9,1.1);
Data(shrinkrows, 'b')=0;

Variable dev, xbase(cols);
Positive Variable xp(cols), xm(cols),xshrink1(cols),xshrink2(cols),xRidge(cols);
Equation eqBase, eqLasso, eqL1Shrink, eqL2Shrink, eqRidgeReg;

eqBase..     dev =e= Sum(rows,    sqr(   Data(rows,'b')-Sum(cols, Data(rows,cols)   *xbase(   cols))))/2;
eqLasso..    dev =e= Sum(rows,    sqr(	 Data(rows,'b')-Sum(cols, Data(rows,cols)   *(xp(cols)-xm(cols)))))/2 
                     + shrinkScale * Sum((cols,shrinkrows), Data(shrinkrows,cols)   *(xp(cols)+xm(cols)));

eqL1Shrink.. dev =e= Sum(rows,    sqr(	 Data(rows,'b')-Sum(cols, Data(rows,cols)   *xshrink1(cols))))/2 
                     + shrinkScale * Sum((cols,shrinkrows), Data(shrinkrows,cols)   *xshrink1(cols));

eqL2Shrink.. dev =e= Sum(allrows, sqr(Data(allrows,'b')-Sum(cols, Data(allrows,cols)*xshrink2(cols))))/2;
eqRidgeReg.. dev =e= Sum(rows,    sqr(Data(rows,'b')-Sum(cols, Data(rows,cols)  * xRidge(cols))))/2 
		     + Sum((cols,ccols), xRidge(cols)*RidgeMatrix(cols,ccols)*xRidge(ccols))/2;

Model modBase    "ols"                                       /eqBase/; 
Model modLasso   "lasso"                                     /eqLasso/; 
Model modShrink1 "non negative lasso model"                  /eqL1Shrink/; 
Model modShrink2 "non negative augmented regression model"   /eqL2Shrink/; 
Model modRidge   "non negative ridge regression"             /eqRidgeReg/; 

Solve modBase    using nlp minimizing dev;
Solve modShrink2 using nlp minimizing dev;
shrinkScale = Sum(shrinkrows, abs(Data(shrinkrows,'b')-Sum(cols, Data(shrinkrows,cols)*xshrink2.l(cols))));
Solve modLasso   using nlp minimizing dev;
Solve modShrink1 using nlp minimizing dev;
Solve modRidge   using nlp minimizing dev;
\end{verbatim} 	
\end{tiny}

%\IfFileExists{./workL1L2.tex}{\clearpage \input{workL1L2.tex} }{}

\end{document}